\documentclass[a4paper]{article}

\usepackage{INTERSPEECH2021}
\usepackage{comment}

\title{Dynamic Encoder Transducer: A Flexible Solution For Trading Off Accuracy For Latency}
\name{Yangyang Shi, Varun Nagaraja, Chunyang Wu, Jay Mahadeokar, Duc Le, Rohit Prabhavalkar, Alex Xiao, Ching-Feng Yeh, Julian Chan, Christian Fuegen, Ozlem Kalinli, Michael L. Seltzer}
\address{
  Facebook AI }
  \email{yyshi@fb.com}

\begin{document}

\maketitle
\begin{abstract}
We propose a dynamic encoder transducer (DET) for on-device speech recognition. One DET model scales to multiple devices with different computation capacities without retraining or finetuning. To trading off accuracy and latency, DET assigns different encoders to decode different parts of an utterance. We apply and compare the layer dropout and the collaborative learning for DET training. The layer dropout method that randomly drops out encoder layers in the training phase, can do on-demand layer dropout in decoding. Collaborative learning jointly trains multiple encoders with different depths in one single model. Experiment results on Librispeech and in-house data show that DET provides a flexible accuracy and latency trade-off. Results on Librispeech show that the full-size encoder in DET relatively reduces the word error rate of the same size baseline by over $8\%$. The lightweight encoder in DET trained with collaborative learning reduces the model size by $25\%$ but still gets similar WER as the full-size baseline. DET gets similar accuracy as a baseline model with better latency on a large in-house data set by assigning a lightweight encoder for the beginning part of one utterance and a full-size encoder for the rest.

\end{abstract}
\noindent\textbf{Index Terms}: Transducer, Transformer, Layer Dropout, Collaborative Learning
\vspace{-6pt}
\section{Introduction}
\vspace{-3pt}
Due to the ubiquity of voice assistants in smartphones, smart speakers, and smart wearable devices, privacy is getting more and more attention. On-device automatic speech recognition (ASR)~\cite{xin_2013_acc,mcgraw_2016_psr,he2019rnnt,jinyu_2019_rnnt} performs the speech recognition entirely on the device. Without streaming audio from the device to the server, on-device ASR provides better privacy protection than server ASR system.

Recurrent neural network transducer (RNN-T)~\cite{Graves2012,he2019rnnt,Rao2017} has been widely applied to on-device ASR. Like many other End-to-end (E2E) models~\cite{Graves2014,ds2,Miao2016,Zhang2019,Shi2019_ICASSP,Das2018}, RNN-T directly optimizes the transduction from the acoustic feature sequence to label sequence by combining acoustic model, pronunciation, and language model all into one neural network. RNN-T intrinsically supports online streaming speech recognition. More importantly, RNN-T models have a much lower footprint than hybrid systems, which is more suitable for on-device ASR. In this work, we used an alignment restricted transducer model~\cite{jay_2020_arrnnt} which gives faster training and better token emission latency.

Recently better accuracy and real-time factors have been achieved by replacing LSTM encoder with transformer~\cite{Vaswani2017} and its variants (e.g. Conformer \cite{conformer}, Emformer \cite{emformer}, etc.) in sequence transducer~\cite{zhang_2020,Yeh_2019,conformer,cfyeh_asru_2020,yongqiang2021_icassp}. In this work, we adopt the Emformer transducer in on-device ASR for low latency speech recognition. The emformer transducer model applies a deep structured transformer as the encoder. Each layer in the encoder contains millions of parameters involving high computation cost and memory consumption. On-device ASR often needs accuracy, computation cost, and latency trade-off to achieve the best user experience for different devices. A high-end device can support a big model to deliver good accuracy and a smooth user experience. In contrast, a low-end device may only support a smaller model with less computation cost and memory consumption. Trade-off sometimes is also needed even in decoding a single utterance. In~\cite{macoskey_2021_bifocal}, inference computation cost reduction can be achieved by pivoting different computation pathways for different parts of the utterance. 

Neural network model pruning~\cite{zhu_2017_pruning,yuan_2019_pruning,han_2015_pruning_quantization},  quantization~\cite{guo_2018_quantization,han_2015_pruning_quantization}, low rank matrix factorization~\cite{dan_2018_tdnnf,xue_2013_svd,ruhit_2016_factorization} and knowledge distillation~\cite{kim_2017_ts,manohar_2019_SLT} have been applied to reduce the footprint and computation cost without significant accuracy loss. However, many of these methods involve finetuning or retraining, which are not flexible. In this work, we propose a dynamic encoder transducer (DET) for on-device ASR with a flexible trade-off between computation cost and accuracy.  

We apply layer dropout and collaborative learning for DET training. Layer dropout~\cite{huang_2016_stochastic_depth} was originally proposed to stabilize and regularize deep convolution neural network training. The work~\cite{pham_2019_ld} applied the layer dropout to train a deep transformer model for speech recognition. The work~\cite{fan_2019_layerdrop} used layer dropout as a structured dropout method to prune the over-parameterized transformer model for many natural language processing tasks. Collaborative learning~\cite{song_2018_cts} leverages the strengths of auxiliary loss, multi-task learning, and knowledge distillation to improve the deep neural classifier's generalization and robustness to label noise. Our work~\cite{varun_2021} applied collaborative learning in transformer transducer to jointly train the teacher and the multiple students all at the same time from scratch. The weight sharing among teachers and students improves the teacher model's performance and all of the student models.

We apply DET in two scenarios for accuracy and computation cost trade-off. The first one is flexibility in tuning encoder depth as a pruning method for a specific device in model deployment.  The other one is dynamically applying different encoders in DET for decoding different parts in one utterance, which is similar to the scenario discussed in ~\cite{macoskey_2021_bifocal,grave_2016_act}. Note the work~\cite{macoskey_2021_bifocal,grave_2016_act} is about LSTM based encoder.

\vspace{-6pt}
\section{Dynamic Encoder Transducer}
\vspace{-3pt}
This section introduces the sequence transducer modeling, the emformer based encoder, the layer dropout method, the collaborative learning method and the DET.

\vspace{-10pt}
\subsection{Sequence transducer}
\vspace{-3pt}
The sequence transducer consists of an encoder, a preditor, and a joiner. Given a sequence of acoustic feature vectors $X=\{x_1, ..., x_T\}$ with length $T$, the encoder $f^e$ generates a sequence of representations $H^e=\{h_1^e,...,h_T^e\}$.
\begin{align}
\vspace{-3pt}
\{h_1^e,...,h_T^e\} = f^e(X). \label{encoder}
\vspace{-3pt}
\end{align}

Let $Y=\{y_1, ..., y_U\}$ in length $U$ where $y_u\in\mathcal{Y}$ be the sequence of output units. We define $\bar{\mathcal{Y}} = \mathcal{Y}\cup\{\phi\}$, where $\phi$ is the blank label. The predictor $f^p$ generates a sequence of representation for the output sequence prefixed with the blank label as follows:
\begin{align}
\vspace{-3pt}
\{h_1^p,...,h_u^p\} = f^p(\{\phi, y_1,...,y_{u-1}\}). \label{predictor}
\vspace{-3pt}
\end{align}

The joiner $f^j$ combines the output representations from the encder and the predictor to generate the logits $h_{t,u}$.
\begin{align}
\vspace{-3pt}
h_{t,u} = f^j(h_t^e, h_u^p). \label{joiner}
\vspace{-3pt}
\end{align}
A softmax function is applied to produce the posterior distribution of the next label $y_u\in\bar{\mathcal{Y}}$
\begin{align}
\vspace{-3pt}
P(y_u|x_{1:t},y_{1:u-1}) = \mathrm{softmax}(h_{t,u}). \label{softmax}
\vspace{-3pt}
\end{align}

Using forward-backward algorithm, the posterior of a sequence of output units $Y=\{y_1, ..., y_U\}$ where $y_u\in\mathcal{Y}$ as follows:
\begin{align}
\vspace{-3pt}
 P(Y|X) = \sum_{A\in\mathcal{B}^-1(Y)} P(A|X), \label{posterior}
 \vspace{-3pt}
\end{align}
where $A$ is one alignment for the sequence of output $Y$. Each label in $A$ is from label set $\bar{\mathcal{Y}}$. $\mathcal{B}$ is the operation to remove the blank from the alignment. The sequence transducer defines the loss as 
\begin{align}
\vspace{-3pt}
L^{Tr} = - log(P(Y|X)). \label{transducer_loss}
\vspace{-3pt}
\end{align}

\vspace{-6pt}
\subsection{Emformer based encoder}
\vspace{-3pt}
We use emformer transducer~\cite{emformer} as a basic model architecture for low latency streaming on-device ASR. The emformer modifies the transformer model with the block processing~\cite{Chunyang_2020_interspeech,dong2019self} to support low latency streaming speech recognition. In training, the emformer uses an attention mask and a ``right context hard copy'' trick to constrain the receptive field for self-attention. In decoding, the emformer applies a cache method to optimize the efficiency optimization by saving the computation for the key and value in self-attention for the left context. 

In the emformer transducer, the encoder accounts for most of the computation cost.  The encoder is much deeper than the predictor (e.g., encoder usually has 10 to 20 layers, predictor only has 2 to 3 layers). More importantly,  the sequence length $T$ in $X=\{x_1, ..., x_T\}$ is much longer than $U$ in $Y=\{y_1, ..., y_U\}$. Reducing the number of layers in the emformer based encoder saves the computation cost. DET focuses on using different encoders with various depths for flexible trade-off accuracy and computation cost in this work.

\vspace{-6pt}
\subsection{Layer dropout}
\vspace{-3pt}
The model becomes robust in decoding with missing layers by randomly masking out some layers in training. The work~\cite{fan_2019_layerdrop} investigated different layer pruning strategies and found that pruning every other layer performed best in inference. 

In this work, we also investigate every other layer of dropout in the training phase. Rather than applying dropout to each layer at random, we apply dropout to every other layer, resulting in a more structured dropout. In decoding, only the layers dropped out in training are pruned for decoding. Every other layer dropout provides less flexibility but makes the training and decoding more consistently.
\begin{figure}[h!]
\vspace{-12pt}
   \begin{center}
    \includegraphics[width=2.5in]{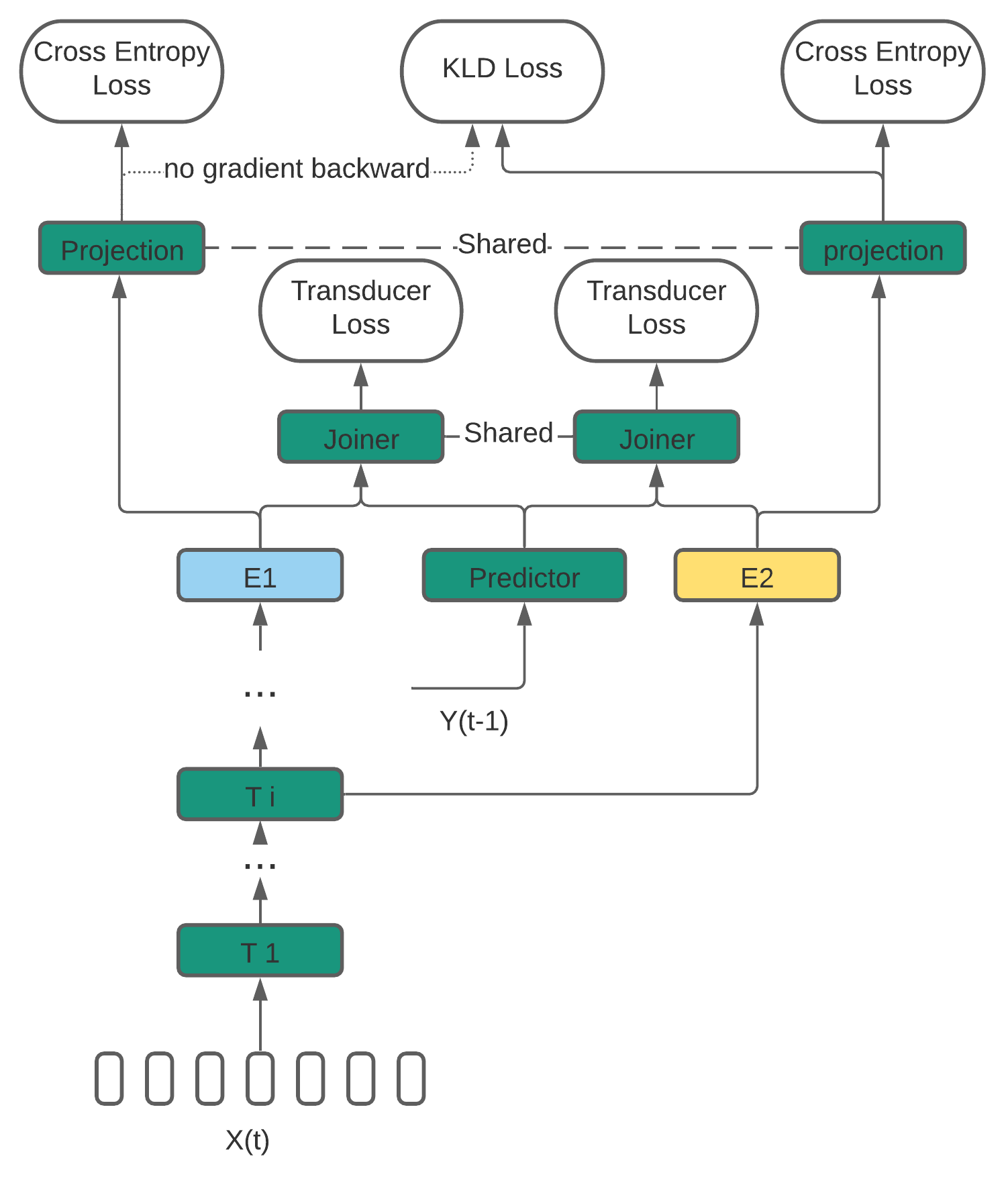}
    \vspace{-8pt}
    \end{center}
    \caption{Collaborative learning for sequence transducer model for on-device ASR. Two encoders with different depths are used. $E_2$ shares all its layers with $E_1$ except the last layer. All the shared nodes by these two encoder are in green.}
    \label{fig:collaborative_learning}
    \vspace{-12pt}
\end{figure}

\vspace{-6pt}
\subsection{Collaborative learning}
\vspace{-3pt}
Figure~\ref{fig:collaborative_learning} gives one DET model with two encoders $E_1$ and $E_2$. The encoder $E_2$ share all its layers with $E_1$ except its last layer. Given $X=\{x_1, ..., x_T\}$, multiple sequence of acoustic representations $H^{i}=\{h_1^{i},...,h_T^{i}\}$ are generated from different encoders $i$. In Figure~\ref{fig:collaborative_learning}, $i\in\{1,2\}$.
\begin{align}
\vspace{-3pt}
\{h_1^{i},...,h_T^{i}\} = f^{i}(X).
\vspace{-3pt}
\end{align}
Combining with the exact predictor's output $\{h_1^p,...,h_u^p\}$, the DET model generates multiple transducer losses $L^{Tr_{i}}$ by passing the different encoders' output through the same joiner.

Similar to the work~\cite{chunxi_2020_aux}, the collaborative learning~\cite{varun_2021} also use the auxiliary supervision based on context-dependent graphemic state (i.e., chenones)~\cite{le2019senones} prediction. Let $C=\{c_1, ..., c_T\}$ be the forced alignment label sequence for $X$. The posterior distribution of $c_t$ from encoder $i$ is obtained as follows:
\begin{align}
\vspace{-3pt}
P^{i}(c_t|x_t) = \mathrm{softmax}(f^{pro}(h_t^{i})),
\vspace{-3pt}
\end{align}
where $f^{pro}$ is a multi-layer perception shared by all encoders. Given the posterior distribution, we can get the sum of the cross-entropy losses for all the encoders. 
\begin{align}
\vspace{-3pt}
L^{CE} = -\frac{1}{T}\sum_{i}\sum_{T}log(P^{i}(c_t|x_t)).\label{CE}
\vspace{-3pt}
\end{align}

The auxiliary cross-entropy loss introduces the forced alignment information into DET training and constructs the bridge to do the knowledge distillation from the full-size encoder $e_1$ to the lightweight encoders $e_i$ where $i > 1$. As shown in Figure~\ref{fig:collaborative_learning}, Kullback–Leibler divergence loss is used. 
\begin{align}
\vspace{-3pt}
L^{KLD} = -\frac{1}{T}\sum_{i>1}\sum_{T}P^{e_{1}}(c_t|x_t)log(\frac{P^{e_{1}}(c_t|x_t)}{P^{i}(c_t|x_t)}).
\vspace{-3pt}
\end{align}
where $e_{1}$ the deepest encoder in DET is used as a teacher for all the rest encoders. Note the gradient from $L^{KLD}$ is not back-propagated to $e_1$. The final loss $L$ to train DET model is 
\begin{align}
L =  \alpha L^{CE} + \beta L^{KLD} + \sum_{i}L^{Tr_{i}}, 
\end{align} where $\alpha$ and $\beta$ are the interpolation weight for the cross-entropy loss and the Kullback–Leibler divergence loss, respectively.

\subsection{Applications of dynamic encoder transducer}
\begin{figure}[h!]
\vspace{-12pt}
   \begin{center}
    \includegraphics[width=3.0in]{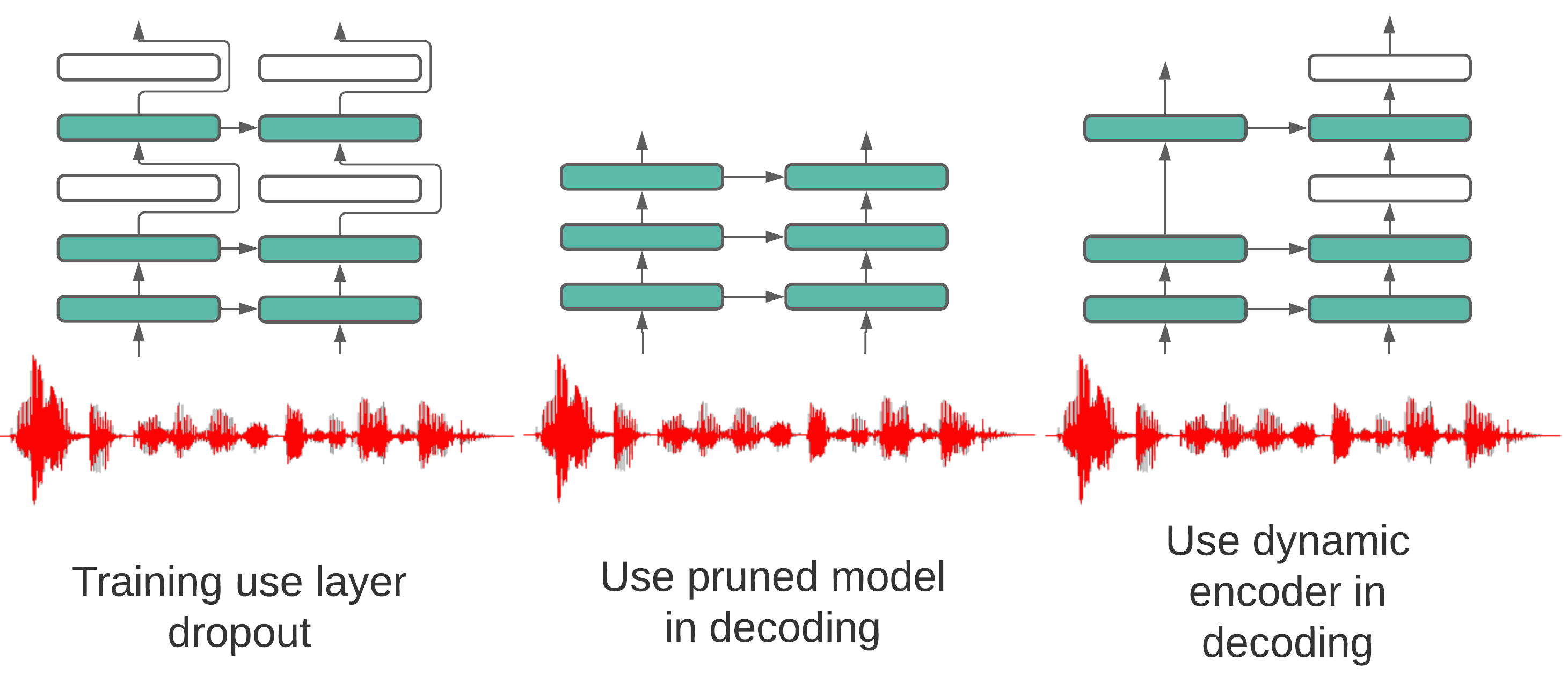}
    \vspace{-8pt}
    \end{center}
    \caption{Two usage scenarios of the DET. The middle part shows that one specific depth encoder is selected. The right part shows that different depth encoders are used for decoding.}
    \label{fig:dynamic_depth_encoder_scenarios}
    \vspace{-12pt}
\end{figure}

Figure~\ref{fig:dynamic_depth_encoder_scenarios} illustrates two application scenarios of the DET model. The figure's middle part shows a typical pruning process where one specific depth encoder is selected from the DET. This process happens when we need to optimize the on-device ASR model computation cost for a specific device. The right part shows the decoding uses two different encoders $e_1$ and $e_2$ jointly to get the representation for acoustic features:
\begin{align}
\vspace{-3pt}
&\{h_1^e,...,h_{T1}^e\} = f^{e_1}(\{x_1,...,x_{K}\}), \\ 
&\{h_{K+1}^e,...,h_{T}^e\} = f^{e_2}(\{x_{K+1},...,x_{T}\}). 
\vspace{-3pt}
\end{align}

Decoding using multiple encoders jointly on one utterance provides a flexible solution to trade-off accuracy and computation cost for a selected device. It is helpful to deal with the one-shot assistant queries with wake words. When a user invokes a voice assistant with a wake word and then speaks a query, the on-device ASR often needs to process the user's query as well as the wake word. Note \textit{audio bursting} happens, where the audio corresponding to the wake word is available for the ASR engine all at once. The user-perceived latency will be improved if a smaller encoder processes the audio bursting with less computation cost.
Additionally, there is a cold-start latency when users start to interact with the voice assistant in a new session. The device needs to load the ASR model into memory. Loading a small encoder initially and gradually loading the rest of the encoder layers in parallel reduces the on-device ASR engine cold-start latency. 

\vspace{-6pt}
\section{Experiments}
\vspace{-3pt}
To evaluate the performance of DET, we carry out two sets of experiments \textit{one encoder decoding} and \textit{dynamic encoder decoding}. \textit{One encoder} simulates the situation that one specific encoder needs to be selected to meet a device's computation capacity. \textit{Dynamic encoder decoding} uses multiple encoders in DET for decoding one utterance.
\vspace{-6pt}
\subsection{Data}
\vspace{-3pt}
\label{subsubsec:training_dataset}
The experiments use LibriSpeech corpus~\cite{librispeech_corpus} and a large in-house dataset. LibriSpeech is an open-source speech corpus that contains 1000 hours of speech derived from audiobooks in the LibriVox project. Approximately 30 hours of data is used for development and evaluation, which are split into \textit{clean} subsets and \textit{other} subsets. 

The in-house dataset consists of data from \textit{Voice Assistant} and \textit{Open Domain}. All data is anonymized with personally identifiable information (PII) removed. The \textit{Voice Assistant} data contains 68k hours of data from  human transcribed data from 20K crowd-sourced workers recorded via mobile devices, 1K hours voice commands, sampled from production traffic and 23k hours of voice command data generated from an in-house TTS system. The \textit{Open Domain} data includes 13K hours of data from public social media English videos that are anonymized with PII removed and annotator transcribed and 1.5M hours from the same source with a large offline model's transcriptions.  

%The \textit{Voice Assistant} data contains 68k hours of data from three parts. The first part is human transcribed data from 20K crowd-sourced workers recorded via mobile devices. The data is further distorted by simulated reverberation and background noise obtained from publicly available videos. The second part is 1K hours voice commands, sampled from production traffic. The data is augmented by speed perturbations~\cite{ko2015audio} at 0.9 and 1.1 times the original speed. The same distortion and additive noise method are applied to the speed perturbed data. The third part is 23k hours of voice command data generated from an in-house TTS system.

%The \textit{Open Domain} data includes two parts. The first part is 13K hours of data from public social media English videos that are anonymized with PII removed and annotator transcribed. We generate a corpus of 78K hours by using the same distortion as above. The second part consists of 1.5M hours from the same source with a large offline model's transcriptions.  

In evaluation, we use \textit{assi} and \textit{dict} dataset. The \textit{assi} is 13.6K manually transcribed de-identified utterances from in-house volunteer employees, which begin with a wake word. The \textit{dict} is 8 hours open domain dictation from crowd-sourced workers recorded via mobile devices. All the evaluation data is anonymized with personally identifiable information (PII) removed. For both Librispeech and in-house dataset, we generate forced alignment using a context and positional dependent graphemes (i.e., chenones)~\cite{le2019senones} based hybrid system.

\vspace{-6pt}
\subsection{Experiment Setting}
\vspace{-3pt}
In all the experiments, we use 80-dimensional log Mel filter bank features at a 10ms frame rate. To increase the training robustness, \emph{SpecAugment} \cite{park2019specaugment} without time warping are used. For the Librispeech experiment, we map the 80-dimensional features by a linear layer to 128 dimension vectors. A 512-dimensional vector is formed by concatenating four continuous 128-dimensional vectors, which is the input to Emformer. For a large in-house dataset, a 480-dimensional superframe is formed by concatenating six continuous features. A linear layer maps the superframe to a 512-dimensional vector.

Each Emformer uses eight heads of self-attention with input dimension 512. The output from the self-attention goes through a feed-forward layer with dimensionality 2048. We set dropout 0.1 for all layers across all experiments. The segment sizes are 160ms in LibriSpeech experiment and 300ms for in-house data experiments. The left context sizes are 1.2s for Librispeech and 1.8s for in-house data. The 40ms and 60ms look-ahead contexts are used for LibriSpeech and in-house experiments, respectively. The 512-dimensional output from the stack of Emformer layers goes through a layer norm followed by a linear layer. Finally, the output from the encoder in the transducer model is a 1024-dimensional vector.

The predictor consists of a 256-dimensional embedding layer, three LSTM layers with 512 hidden nodes, and a linear projection layer with 1024 output nodes. The combined 1024 dimensional embeddings from the encoder and the predictor go through a Tanh activation and then another linear projection to the final output with 4096 sentence pieces~\cite{kudo2018sentencepiece}. 

All models are trained with the adam optimizer~\cite{kingma2014adam} using warming-up updates. The learning rate is 1e-3 for all the experiments. In Librispeech experiments, the last checkpoint from 120 epoch training is for evaluation. For large in-house data, we use the checkpoint from 800K updates for evaluation. In collaborative training, we set both $\alpha$ and $\beta$ to 0.5. In layer dropout, the optimal dropout rate for each layer is 0.1 obtained by grid search. 

All the model training uses 32 Nvidia V100 GPUs. We evaluate the latency by the real-time factors (RTFs) and speech engine perceived latency (SPL). SPL measures the time from speech engine gets the last word from user utterance to speech engine transcribes the last word and gets the endpoint signals. For in-house data experiment, the SPL evaluation uses the in-house static endpointer, neural endpointer, and transducer's end-of-sequencing symbol. We use low-end android device for latency evaluation. We sample 100 utterances from \textit{clean} and \textit{assi} for latency evaluation for Librispeech and in-house data, respectively.
\begin{table}[htbp]
    \centering
    \begin{tabular}{|cc|cc|c|}
    \hline
    arch &\# layers &\textit{clean} & \textit{other} & $\#$params \\
    \hline\hline
     base &20 &  3.62 & 9.86  & 77M  \\
     base pruned &14 &  31.42 & 49.50  & 58M  \\
     base &14 & 3.87 & 10.35 & 58M  \\
     \hline
    random &20 &  3.75 & 9.38 & 77M \\
    6-16:2 &14 & 4.80 & 12.45& 58M \\ 
    1-16:3 &14 & 4.35 & 11.56& 58M  \\ 
    \hline 
    group  & 20&3.97 & 10.09 & 77M\\
    1-16:3 & 14 & 4.32 & 11.38& 58M\\
    \hline 
    cl &20 & \textbf{3.54} & \textbf{9.04} &77M\\
    cl &14 & 3.66 & \textbf{9.60}& 58M\\
    \hline 
    \end{tabular}
    \caption{WER and RTFs on LibriSpeech data for DET. ``base pruned'' directly prunes the full size encoder. ``random'' dropout each layer randomly. ``group'' dropout layer 1, 4, 7, 10, 13 and 16 as a group randomly. ``6-16:2'' dropouts every 2 layers from layer 6 to layer 16. ``1-16:3'' dropouts every 3 layers from layer 1 to layer 16. ``cl'' denotes the DET trained using collaborative learning.}
    \label{tab:one_encoder_libri}
    \vspace{-18pt}
\end{table}

\begin{table}[htbp]
\vspace{-6pt}
    \centering
    \begin{tabular}{|cc|cc|c|c|}
    \hline
    arch & $\#$layers &\textit{dict} & \textit{assi} & RTFs &SPL\\
    \hline\hline
     baseline &20 &  16.40 & 3.83  & 0.47 & 756 \\
     baseline &14 & 17.56 & 4.04  & 0.40 & 697 \\
     \hline
     random &20 &  16.21 & 3.95 &0.47 &766 \\
     1-16:3 &14 & 19.29 & 6.74 & 0.39 &688\\
    \hline 
    cl &20 & 17.41 & 4.01 & 0.45 &732\\
    cl &14 & 18.87 & 4.53 & 0.38 &685 \\
    \hline 
    \end{tabular}
    \caption{WER, RTFs and SPL on in-house data for DET trained by layer dropout and collaborative learning. ``random'' dropout each layer randomly. ``1-16:3'' dropouts every 3 layers from layer 1 to layer 16. ``cl'' denotes the DET trained using collaborative learning.}
    \label{tab:one_encode_inhouse}
    \vspace{-20pt}
\end{table}

\vspace{-6pt}
\subsection{One encoder decoding}
\vspace{-3pt}
In one encoder decoding, both layer dropout and collaborative learning are used as prune methods to select one encoder that meets one specific device's requirement. Table~\ref{tab:one_encoder_libri} and Table~\ref{tab:one_encode_inhouse} gives the word error rate (WER) for DET on Librispeech data and in-house data, respectively. In Table~\ref{tab:one_encoder_libri}, we can see that the 20-layers encoder trained by the layer dropout and collaborative learning reduces the WER of the same size baseline on \textit{other} by $5\%$ and $8\%$, respectively. ``base pruned'' shows that directly pruning the 20 layers baseline model to 14 layers does not work. In Table~\ref{tab:one_encoder_libri}, the best-pruned model with 14 layers is from collaborative learning, which gets the on-par accuracy as the full-size model. The pruned models from both the ``random'' and ``group'' dropout method are worse than re-training the same size model from scratch. Note the results from collaborative learning in Table~\ref{tab:one_encoder_libri} is different with~\cite{varun_2021}, in this paper, each student encoder shares its layers except the last one.

Table~\ref{tab:one_encode_inhouse} shows that the pruned model trained by collaborative learning is more accurate than the one trained from layer dropout. However, neither of the pruned models outperforms the same size baseline model trained separately. For in-house data, the full-size encoder trained by layer dropout performs similarly as same size baseline with a slight improvement on \textit{dict} and slight degradation on \textit{assi}.

\vspace{-6pt}
\subsection{Dynamic encoder in decoding}
\vspace{-3pt}
Table~\ref{tab:lib_det} and Table~\ref{tab:inhouse_det} show the experiment results using multiple encoders in decoding one utterance on Librispeech data and in-house data, respectively. For Librispeech data, we compare the layer dropout and the collaborative learning. For in-house data, we only use the layer dropout method. The full-size encoder in DET trained by the layer dropout is more accurate than the same-size encoder trained by collaborative learning.

Both tables show that by adjusting the audio's length to be decoded by the small encoder in DET, we can do a trade-off between accuracy and latency. Table~\ref{tab:lib_det} shows that decoding using DET trained by collaborative learning gives better WER and RTFs trade-off than layer dropout. Using a small encoder for the beginning 0.8s audio, the DET gets $6\%$ relative WER reduction and over $10\%$ RTFs reduction.

\begin{table}[htbp]
    \centering
    \begin{tabular}{|cc|cc|c|c|}
    \hline
    arch & time &\textit{clean} & \textit{other} & $\#$layers & RTFs\\
    \hline\hline
     base &-&  3.62 & 9.86&20   & 0.63\\
     base &-& 3.87 & 10.35&14  &0.47 \\
     \hline
     drop   &6.4s&4.18  &11.07 & 14/20 & 0.51\\ 
     drop    &3.2s & 4.04&10.45 & 14/20 & 0.52 \\ 
     drop   &1.6s&3.90  &10.02 & 14/20 & 0.57 \\ 
     drop    &0.8s & 3.84 &9.49 & 14/20 & 0.60 \\ 
    \hline 
    cl  & 6.4s& 3.75& 9.63 & 14/20&0.51\\
    cl  & 3.2s &3.76 &9.52 & 14/20&0.54\\
    cl  & 1.6s& 3.72& 9.45 & 14/20&0.55\\
    cl  & 0.8s &\textbf{3.62} &\textbf{9.27} & 14/20&0.56\\
    \hline 
    \end{tabular}
    \caption{WER and RTFs on LibriSpeech data for DET trained by layer dropout and collaborative learning. ``time'' column denotes the length of the beginning part of audio decoded by the small encoder. ``drop'' and ``cl'' denotes the DET trained using layer dropout and collaborative learning, respectively.}
    \label{tab:lib_det}
    \vspace{-20pt}
\end{table}

Table~\ref{tab:inhouse_det} shows that using multiple encoders for one utterance decoding achieves advantages from each encoder. In the experiment, we use multiple encoders decoding for \textit{assi} which has a wake word at the beginning for each utterance. We use the full-size encoder for \textit{dict} which does not have wake word. The bottom line in Table\ref{tab:inhouse_det} shows that using a small encoder for the beginning 1.5s audio gives slight WER degradation with better RTFs and SPL. 

\begin{table}[htbp]
    \centering
    \begin{tabular}{|ccc|cc|c|c|}
    \hline
    arch & time &$\#$layers &\textit{dict} & \textit{assi} & RTFs &SPL\\
    \hline\hline
     baseline& - &20 &  16.40 & 3.83  & 0.47 & 756 \\
     baseline& - &14 & 17.56 & 4.04  & 0.40 & 697 \\
    \hline 
    drop & 3.0s&14/20& 16.21&4.04 & 0.41 & 700 \\
    drop & 1.5s&14/20& 16.21&3.99 & 0.44 & 714 \\
    \hline 
    \end{tabular}
    \caption{WER, RTFs and SPL on in-house data for DET trained by layer dropout.  ``1-16:3'' dropouts every 3 layers from layer 1 to layer 16. ``drop'' denotes the DET trained using layer dropout. ``time'' column denotes the length of the beginning part of audio decoded by the small encoder.}
    \label{tab:inhouse_det}
    \vspace{-20pt}
\end{table}

\vspace{-12pt}
\section{Conclusions}
\vspace{-6pt}
This paper proposed a dynamic encoder transducer as a flexible on-device ASR model for accuracy and latency trade-off. Two application scenarios were discussed: pruning encoder to fit device computation limit and decoding different parts of one utterance using different encoders. We applied layer dropout and collaborative learning for DET model training. The layer dropout method that randomly masked some encoder layers in training makes DET flexible and robust in dropping layers in decoding. The collaborative learning jointly trained multiple encoders with different depths using weight sharing, auxiliary tasks, and knowledge distillation. On Librispeech data, experiments showed that the DET model outperformed the baseline in accuracy and latency for both scenarios. In-house data experiments showed that DET gets on par accuracy as baseline model but with better RTFs and latency. 

\bibliographystyle{IEEEtran}
\newpage
\bibliography{mybib}

% \begin{thebibliography}{9}
% \bibitem[1]{Davis80-COP}
%   S.\ B.\ Davis and P.\ Mermelstein,
%   ``Comparison of parametric representation for monosyllabic word recognition in continuously spoken sentences,''
%   \textit{IEEE Transactions on Acoustics, Speech and Signal Processing}, vol.~28, no.~4, pp.~357--366, 1980.
% \bibitem[2]{Rabiner89-ATO}
%   L.\ R.\ Rabiner,
%   ``A tutorial on hidden Markov models and selected applications in speech recognition,''
%   \textit{Proceedings of the IEEE}, vol.~77, no.~2, pp.~257-286, 1989.
% \bibitem[3]{Hastie09-TEO}
%   T.\ Hastie, R.\ Tibshirani, and J.\ Friedman,
%   \textit{The Elements of Statistical Learning -- Data Mining, Inference, and Prediction}.
%   New York: Springer, 2009.
% \bibitem[4]{YourName17-XXX}
%   F.\ Lastname1, F.\ Lastname2, and F.\ Lastname3,
%   ``Title of your INTERSPEECH 2021 publication,''
%   in \textit{Interspeech 2021 -- 20\textsuperscript{th} Annual Conference of the International Speech Communication Association, September 15-19, Graz, Austria, Proceedings, Proceedings}, 2020, pp.~100--104.
% \end{thebibliography}

\end{document}